\documentclass[conference]{IEEEtran}
\IEEEoverridecommandlockouts
\usepackage{cite}
\usepackage{amsmath,amssymb,amsfonts}
\usepackage{graphicx}
\usepackage{textcomp}
\usepackage{xcolor}
\def\BibTeX{{\rm B\kern-.05em{\sc i\kern-.025em b}\kern-.08em
    T\kern-.1667em\lower.7ex\hbox{E}\kern-.125emX}}

\usepackage{amsmath,amsfonts,bm}

\def\eqref#1{equation~\ref{#1}}

\def\1{\bm{1}}

\def\vmu{{\bm{\mu}}}

\def\vx{{\bm{x}}}

\def\vz{{\bm{z}}}

\def\mD{{\bm{D}}}
\def\mE{{\bm{E}}}

\def\mI{{\bm{I}}}

\def\mL{{\bm{L}}}
\def\mM{{\bm{M}}}

\def\mO{{\bm{O}}}
\def\mP{{\bm{P}}}

\def\mLambda{{\bm{\Lambda}}}
\def\mSigma{{\bm{\Sigma}}}

\DeclareMathAlphabet{\mathsfit}{\encodingdefault}{\sfdefault}{m}{sl}
\SetMathAlphabet{\mathsfit}{bold}{\encodingdefault}{\sfdefault}{bx}{n}

\newcommand{\R}{\mathbb{R}}

\usepackage{amsmath}
\usepackage{amssymb}
\usepackage{hyperref}
\usepackage{cleveref}
\usepackage{bm}
\usepackage[acronym,nomain,shortcuts]{glossaries}
\usepackage{url}
\usepackage{layouts}
\usepackage{enumitem}
\usepackage{diagbox}
\usepackage[export]{adjustbox}
\usepackage{multirow}
\usepackage{hhline}
\usepackage{booktabs}
\usepackage{makecell}
\usepackage{float}
\glsdisablehyper
\usepackage{stfloats} 
    
\newcommand{\enquote}[1]{"#1"}
\newcommand{\mGamma}{{\bm{\Gamma}}}

\newcommand{\vxi}{{\bm{\xi}}}
\newcommand{\tvx}{\tilde{\vx}}

\newcommand{\vepsilon}{{\bm{\epsilon}}}

\newcommand{\vGamma}{{\bm{\Gamma}}}
\let\oldsqrt\sqrt
\def\sqrt{\mathpalette\DHLhksqrt}
\def\DHLhksqrt#1#2{%
	\setbox0=\hbox{$#1\oldsqrt{#2\,}$}\dimen0=\ht0
	\advance\dimen0-0.2\ht0
	\setbox2=\hbox{\vrule height\ht0 depth -\dimen0}%
	{\box0\lower0.4pt\box2}}

\crefname{section}{Sec.}{Secs.}
\crefname{subsection}{Sec.}{Secs.}
\crefname{subsubsection}{Sec.}{Secs.}
\crefname{paragraph}{Par.}{Para.}
\crefname{appendix}{App.}{Apps.}
\crefname{table}{Tab.}{Tabs.}
\crefname{figure}{Fig.}{Figs.}
\crefname{equation}{Eq.}{Eq.}
\Crefname{equation}{Eq.}{Eq.}
\crefname{algorithm}{Alg.}{Algs.}
    
\begin{document}

\title{Large-scale gradient-based training of Mixtures of Factor Analyzers\\
}

\author{\IEEEauthorblockN{Alexander Gepperth}
	\IEEEauthorblockA{\textit{Fulda University of Applied Sciences}\\
		Fulda, Germany \\
		alexander.gepperth@cs.hs-fulda.de}
}

\maketitle

\begin{abstract}
Gaussian Mixture Models (GMMs) are a standard tool in data analysis. However, they face problems when applied to 
high-dimensional data (e.g., images) due to the size of the required full covariance matrices (CMs), whereas the use of 
diagonal or spherical CMs often imposes restrictions that are too severe.
The Mixture of Factor analyzers (MFA) model is an important extension of GMMs, which allows to smoothly interpolate between 
diagonal and full CMs based on the number of \textit{factor loadings} $l$. MFA has successfully been applied for modeling high-dimensional image data \cite{Richardson2018}.
This article contributes both a theoretical analysis as well as a new method for efficient high-dimensional MFA training by stochastic gradient descent, starting from random centroid initializations. This greatly simplifies the training and initialization process, and avoids problems of batch-type algorithms such Expectation-Maximization (EM) when training with huge amounts of data.
In addition, by exploiting the properties of the matrix determinant lemma, 
we prove that MFA training and inference/sampling can be performed based on precision matrices, which does not require matrix inversions after training is completed. 
At training time, the methods requires 
the inversion of $l\times l$ matrices only.
%
Besides the theoretical analysis and proofs, we apply MFA to typical image datasets such as SVHN and MNIST, and demonstrate the ability to perform sample generation and outlier detection.
\end{abstract}

\begin{IEEEkeywords}
Gaussian Mixture Models, Stochastic Gradient Descent, Mixture of Factor Analyzers, Mixture Models
\end{IEEEkeywords}

\section{Introduction}
This contribution focuses on generative machine learning models applied to images. In the original sense of the term (see \cite{Bishop2006pattern}), this implies that a learning algorithms aims at explicitly modeling the distribution of the data. This includes sampling from this distribution, but is by no means limited to this function. Most prominently, the direct modeling of the distribution itself allows outlier detection and inference.

Mixture models figure prominently among generative models, since their approach is to represent the overall data distribution as a linear combination of elementary distributions $\Psi_k(\vx)$, each of which has parameters $\vxi_k$:
\begin{align}
p(vec x) = \sum_k \pi_k \Psi(\vx ; \vxi_k) \equiv \sum_k \pi_k \Psi_k(\vx)
\end{align}
Mixture modeling allows the derivation of important analytical results, e.g., if the $\Psi_k$ are mathematically well-understood. 
A particular case is the use of multi-variate normal distributions:
\begin{align}
\Psi_k(\vx) = \mathcal N(\vx ; \vmu_k,\mSigma_k) \equiv \mathcal N_k(\vx),
\end{align}
which gives rise to so-called \textit{Gaussian Mixture Models)} (GMMs).

MFA represent a special case of GMMs: for describing $d$-dimensional variables, MFA assumes the existence of an $l$-dimensional \textit{latent space} with $l$\,$\ll$\,$d$ for each mixture component $k$.
In the latent space, variables $\R^l$\,$\in$\,$\vz_k$ follow a simple normal distribution: $\vz_k$\,$\sim$\,$\mathcal{N}(0,\mI\in\R^{l\times l})$. The relation between latent and full space is given by a simple generative model:
\begin{equation}\label{eq:gen}
\R^d\ni\vx_k = \vmu_k + \mLambda_k \vz_k + \vepsilon_k
\end{equation}
with $\vepsilon_k$\,$\sim$\,$\mathcal{N}(0,\mD_k)$, $\mD_k \in \R^{d\times d}$ diagonal and $\mLambda_k\in\R^{d\times l}$.
From this, it directly follows that $\vx_k$\,$\sim$\,$\mathcal{N}(\vmu_k,\mSigma_k)$ with 
\begin{equation}
\mSigma_k \equiv \mD_k + \mLambda_k\mLambda_k^T.
\label{eq:cov1}
\end{equation}. 
Inversely, the probability that a sample $\vx$ was generated by mixture component $k$ is given as:
\begin{equation}\label{eq:cov}
\ln p_k(\vx) = -\frac{1}{2}\Big\{ d\log(2\pi) + \log \det \mSigma_k + \tvx_k^T\mSigma_k^{-1}\tvx_k \Big\}.
\end{equation} 
We will use the convenient shorthand $\tvx_k\equiv\vx - \vmu_k$ throughout this article.
The goal of {MFA} training is to find the \textit{loading matrices} $\mLambda_k$, the component noise matrices $\mD_k$, the component means $\vmu_k$ and the component weights $\pi_k$. This is achieved by maximizing the MFA log-likelihood:
\begin{align}\label{eqn:ll}
\mathcal L = \sum_n \log\sum_k p_k(\vx_n)
\end{align}
Optimization of $\mathcal L$ is typically done using the Expectation-Maximization (EM) algorithm which is applicable to many latent-variable models. It relies on the iterative optimization
of a lower-bound to the model log-likelihood, which can be shown to be tight.
Since EM is not conceptually based on gradient descent, it does not require 
learning rates or step sizes to be set and is thus very easy to handle. 
%
\subsection{Motivation}
While EM training is feasible for small amounts of low-dimensional data, SGD is a preferable alternative in other cases:
\\
\par\noindent\textbf{Efficient training for large-scale problems} 
When the number of training samples is high, 
EM as a batch-type algorithm becomes infeasible since it requires a pass through the whole dataset for each iteration. Stochastic generalizations of EM to mini-batch type learning exist \cite{cappe2009line}, but involve several new hard-to-tune hyper-parameters 
Here, SGD offers a principled alternative. Being based on mini-batches and having a solid theoretical foundation in the Robins-Monro procedure \cite{monro}, it can be applied to arbitrarily large datasets. 
\\
\par\noindent\textbf{Training from random initial conditions}
EM convergence is strongly dependent on initialization, and thus initializing EM by a clustering algorithm such as k-means is common. This, too, becomes problematic for large-scale problems since k-means is a batch-type algorithm as well. We therefore 
propose training MFA by SGD from random initial conditions, which greatly simplifies 
the training procedure and removes the necessity to process the whole dataset for initialization.
\\
\par\noindent\textbf{Efficiency of MFA for high-dimensional data} 
MFA training involves the inversion of large matrices for high latent-space dimensionalities. since the log-likelihood \ref{eq:cov} contains a determinant. However, in \ref{eq:cov}, covariance matrices must be inverted even when no training is performed. Formulating MFA in terms of precision matrices (i.e., inverse to covariance matrices) will remove this requirement.
\subsection{Related Work}
The MFA model was introduced in \cite{Ghahramani1997}, together with the basic idea of using the Woodbury matrix inversion theorem and the matrix determinant lemma to avoid inverting large matrices during MFA training. 
Application of MFA to high-dimensional (image) data was proposed by \cite{McLachlan2003,McLachlan2005}, using the same mathematical ideas for ensuring efficiency even for high data dimensions. In addition to treating high-dimensional image data, 
\cite{Richardson2018} proposes training MFA on large-scale datasets like the CelebA face database and demonstrates excellent image generation performance. In this article, stochastic gradient descent is used instead of EM due to the large amount of samples, although the model is still needs to be initialized by k-means.

Extending MFA to a deep model was first proposed by \cite{Tang2012}, where one MFA instance was trained on the latent variables extracted from data by another one. Advantages of this approach when training on high-dimensional data are described, although it is mentioned that more than two layers are rarely useful. This idea is expanded upon by \cite{Viroli2019}, which introduces deeper MFA models based on a similar principle, although application to image data is not described.
Both \cite{Tang2012} and \cite{Viroli2019} exclusively use batch-type EM for training.

All described works on MFA employ the variance-based description of MFA, meaning that the quantities to be optimized are covariance matrices as opposed to precisions.
\subsection{Goals and contributions}
The goals of this article are to establish MFA training by SGD as a simple and scalable alternative to EM and sEM in streaming scenarios with potentially high-dimensional data. 
For doing so, we build upon previous work on the optimization of GMMs by SGD \cite{gepperth2021c}. 
The main novel contributions are:

\begin{itemize}[leftmargin=*,nosep]
	\item Mathematical analysis to prove that optimization of MFA log-likelihood by SGD is feasible
	\item Proof that MFA can be formulated in terms of precision matrices
	\item Procedure to train MFA by SGD from random initial conditions
	\item Demonstration of MFA sampling and outlier detection on typical image datasets (MNIST, SVHN)
\end{itemize}
Additionally, we provide a TensorFlow implementation.\footnote{\href{https://github.com/anon-scientist/mfa}{https://github.com/anon-scientist/mfa}} %
\section{Data}
We use the following datasets:
\par
\textbf{MNIST}~\cite{LeCun1998} is the common benchmark for computer vision systems and classification problems.
It consists of $60\,000$ $28$\,$\times$\,$28$ gray scale images of handwritten digits (0-9). \\
\textbf{FashionMNIST}~\cite{Xiao2017} consists of images of clothes in 10 categories and is structured like the MNIST. 
It should be more challenging than the MNIST dataset.\\
%
%
%
\section{Mathematical Analysis}\label{sec:math}
\subsection{Efficient formulation of MFA}
As shown in \cite{Ghahramani1997,Richardson2018}, the component log-likelihoods 
can be expressed without having to store and invert $d$-dimensional matrices by exploiting the Woodbury matrix inversion lemma and the matrix determinant lemma:
\begin{align}
\log \det \mSigma_k &= \log\det \mL_k + \log\det \mD_k \label{eqn:trick}\\
\mSigma_k^{-1}      &= (\mLambda_k \mLambda_k^T +  \mD_k)^{-1} \nonumber\\
                    &= \mD_k^{-1} - \mD_k^{-1}\mLambda_k\mL^{-1}_k \mLambda_k^T\mD_k^{-1}\nonumber
\end{align}
where $\mL_k$\,$\equiv$\,$\mI$\,$+$\,$\mLambda_k^T \mD_k^{-1}\mLambda_k$ is an $l$\,$\times$\,$l$ matrix and thus very efficient to compute and process. 
\subsection{Proof that MFA can be performed based on precisions}\label{sec:prec}
\Cref{eq:cov} could be directly used for gradient descent, which is however difficult in practice due to the explicit matrix inversions. Although the computational load of the matrix inversions in \cref{eqn:trick} is reduced w.r.t. inverting full covariance matrices, is still significant and grows as $\mathcal{O}(l^3)$. Worse still, matrix inversion may be numerically problematic in SGD for initial solutions that are far from the optimum (unlike \cite{Richardson2018} which performs SGD starting from a k-means initialization). Lastly, \cref{eq:cov} requires matrix inversions not only for training, but even when executing a trained model.
\par
We therefore resort to a more efficient and numerically robust strategy which avoids explicit matrix inversions.
To this effect, we parameterize multi-variate normal densities by precision matrices $\mP_k\equiv \mSigma^{-1}_k$. 
While the parameters of the generative model \cref{eq:gen} have a simple relation to the learned covariance matrices, the same is not generally true 
for precision matrices. The question is thus how the parameters $\mLambda_k$, $\mD_k$ of the generative model \cref{eq:gen} can be recovered from a trained precision matrix.

For achieving this, we parameterize precision matrices differently from \cref{eq:cov1}, using diagonal $d\times d$ matrices $\mE_k$ and $d\times l$ matrices $\mGamma_k$ as 
\begin{equation}
\mP_k = \mSigma_k^{-1} = \mE_k - \mGamma_k \mGamma_k^T.
\label{eq:precrepr}
\end{equation}

With this definition, the log-likelihoods can be computed in a more efficient fashion than it would be possible when using covariance matrices, compare to \cref{eq:cov}:
\begin{align}
\log p_k(\vx) &= -\frac{1}{2}\Big{\{} d\log(2\pi) - \log \det \mP_k + \tvx^T\mP_k\tvx \Big\} \nonumber \\
              &= -\frac{1}{2}\Big{\{} d\log(2\pi) - \log \det \mP_k + \tvx^TE\tvx -\nonumber\\
                               & - (\Gamma_k^T\tvx)^2\Big\}
\label{eq:llprec}
\end{align} 
By the matrix determinant lemma, we again have
\begin{align}
\log \det \mP_k &= \log\det \mM_k + \log\det \mE_k \label{eqn:precdet} 
\end{align}\label{eqn:precdet}
where $\mM_k$\,$\equiv$\,$\mI$\,$-$\,$\mGamma^T_k \mE_k^{-1}\mGamma_k$ is again an $l$\,$\times$\,$l$ matrix.

The minus sign in \cref{eq:precrepr} ensures that, by the Woodbury matrix inversion lemma, we can express the covariance matrices $\mSigma_k$ as a function of $\mE_k$ and $\mGamma_k$, which in turns allows the determination of the generative model parameters:
\begin{equation}
\mSigma_k = \mP_k^{-1} = \left( \mE_k-\mGamma_k\mGamma_k^T\right)^{-1} = \mE_k^{-1}+\mE_k^{-1}\mGamma_k\mM^{-1}_k\mGamma_k^T\mE_k^{-1}
\label{eq:inv}
\end{equation}
Examining \cref{eq:inv} and remembering that, 
as a consequence of the formulation of MFA as a generative model, we have $\mSigma_k = \mD_k + \mLambda_k\mLambda_k^T$, we can compare terms and arrive at the following identifications:
\begin {align}
 \mD_k & \equiv \mE_k^{-1} \nonumber\\
\mLambda_k & \equiv\mE_k^{-1}\mGamma_k\mathcal{M}^{-0.5}_k\mO_k \nonumber\\
\mO_k\mathcal{M}^{-1}\mO_k^T & = \mM^{-1}_k\label{eqn:id}
\end{align}
The \enquote{square root} of the inverse of $\mM_k$ is performed by first diagonalizing $\mM_k$ as
$\mM_k = \mO^T \mathcal{M}_k\mO$, with $\mO$ orthogonal and $\mathcal{M}_k$ diagonal. The inverse of the diagonal matrix $\mathcal{M}_k$ is trivial, and we
obtain $\mM_k^{-1} = \mO_k^T\mathcal{M}_k^{-1/2}\mathcal{M}_k^{-1/2}\mO_k$. The orthogonal matrix $\mO_k$ can actually be omitted from $\mLambda_k$ here because an orthogonal transformation, when applied to the random latent vector, will produce another random vector of unit variance and zero mean.
This defines all required parameters of the generative model in terms of the learned precision matrices, which allows us to sample efficiently in the low-dimensional space. 
\subsection[ALT]{Properties of $\mM_k$}\label{sec:m}
The following properties of the $l\times l-$matrices $\mM_k$ are relevant to SGD optimization. Here, we will just present some proofs and comments on why certain properties are desirable:\\

\par\noindent\textbf{Symmetry}:$\mM_k^T = \mM_k$. This automatically follows from the definition of $\mM_k$:
\begin{align}
\mM_k^T &= (I-\mGamma_k \mE_k^{-1}\mGamma_k^T)^T\nonumber\\
        &= I-(\mGamma_k^T)^T(\mE_k^{-1})^T\mGamma_k^T\nonumber\\
        &= I-\mGamma_k \mE_k^{-1}\mGamma_k^T 
         = \mM_k
\end{align}
since the inverse of the diagonal matrix $\mE_k$ is diagonal as well and thus symmetric. Symmetry is important since the eigenvalues of a symmetric real matrix are real, and, by positive-definiteness, must be strictly positive. As a consequence, the definiteness of the $\mM_k$ can be monitored via their eigenvalues. These are cheap to compute since $l$ is usually small.
\\
\par\noindent\textbf{Diagonality} This is not strictly required but facilitates the computation of eigenvalues and, above all, ensures linear independency of the columns of the loading matrices $\mGamma_k$, see \cref{sec:constr}. This can be proven as follows: for any two columns $\vGamma_{k:i}$, $\vGamma_{k:j}$ of the loading matrices, we know that $\mM_{ij} = \mI_{ij}-\vGamma_{k:i}^T\mE^{-1}\vGamma_{k:j}$. Diagonality of $\mM$ implies that $\mM_{ij} = 0$ for $i\neq j$. 
If, on the other hand,  $\vGamma_{k:i}$ and $\vGamma_{k:j}$, $i\neq j$ were linearly dependent ($\vGamma_{k:i}=k\vGamma_{k:j}$, $k\in\mathbb R$), we would have $\mM_{ij}$=$\mI_{ij}-k^2\vGamma_{k:i}^T\mE_k^{-1}\vGamma_{k:i}$ which cannot be zero due to the positive-definiteness of $\mE_k$. Thus, it is shown that diagonalizing the $\mM_k$ leads to linear independency of the columns in the loading matrices. 
\\
\par\noindent\textbf{Positive-definiteness} 
The matrices $\mM_k = \mI - \mGamma_k^T \mE_k^{-1}\mGamma_k$ are not positive-definite by construction, although this property is a requirement of the model. Since we showed in the previous paragraph that 
linearly independent columns of the $\mGamma_k$ achieve diagonality of the $\mM_k$, all we need to show is that there exist choices of particular $\mGamma_k$ and $\mE_k$ for which one diagonal element of $\mM_k < 0$. 
Such a choice is, e.g.: $(\mGamma_k)_{1,:} = [2,0,\dots]^T$ and $\mE_k = I$. In this case, we have 
$M_{11} = -3$ which proves our proposition. The remaining columns of $\mGamma_k$ are arbitrary as long as they are independent from the first one. In order to maintain positive-definiteness, the eigenvalues of the diagonal $\mM_k$ must therefore be monitored and the appropriate columns in $\mGamma_k$ modified in case eigenvalues drop below 0.
\subsection{Sampling from a trained MFA model}\label{sec:sampling}
Sampling is now a straightforward procedure which is conducted in two steps: initially, a component $k^*$ is selected by drawing from a multinomial distribution parameterized by the component weights $\pi_k$: $k^* \sim M(\pi_1, \dots, \pi_K)$. For the selected component $k^*$, 
a realization of the latent variable $\vz$\,$\sim$\,$\mathcal{N}(0,\mI)$ is drawn and then transformed to the high-dimensional space using the generative model of \cref{eq:gen}:
\begin{align}
\vx_{k^*} &= \vmu_{k^*}+\mLambda_{k^*} \vz + \vepsilon_{k^*}\nonumber\\
\vepsilon_{k^*} &\sim \mathcal{N}(0,\mD_{k^*}).
\end{align} 
Here, we compute the loading matrices $\mLambda_{k^*}$ and the diagonal covariance matrices $\mD_{k^*}$ appearing in the generative model from the results of 
precision-based training as indicated in \cref{eqn:id}. As stated there, we omit orthogonal matrices from the definition of the loading matrices, and finally use:
\begin{align}
\mLambda_{k^*} & \equiv\mE_{k^*}^{-1}\mGamma_{k^*}\mathcal{M}^{-0.5}_{k^*}
\end{align}

This two-step sampling procedure is equivalent to that of a GMM, except for the fact that the full-rank covariance matrices in GMMs are here expressed by the loading matrices of lower rank: 
$\mSigma_{k^*} = \mLambda_{k^*}\mLambda_{k^*}^T + \mD_{k^*}$.
This is a general property of MFA models, irrespectively of the way they are trained.
\section{Proposal for constrained SGD optimization}
\label{sec:constr}
When optimizing the precision-based MFA model by gradient descent, we use the procedure for training GMMs by SGD as described in \cite{gepperth2021c}. In addition, several MFA-specific constraints are enforced in addition to the usual GMM constraint $\sum_k \pi_k=1$:\\

\par\noindent\textbf{Linearly independent columns of the loading matrices $\mGamma_k$} 
Although it is not formally required by the model, the columns of the $\mGamma_k$ should be at least linearly independent. If they were not, the dependent columns 
would not contain additional information about the data and could thus be discarded. Indeed, it can be shown that a solution where two columns of the loading matrix are dependent or identical constitutes a local extremal point of the loss (to be avoided for SGD). 
For addressing the independence constraints, we first observe that, if the diagonal precision matrices are spherical, $\mE_k = c\mI$, 
the diagonalization of $\mM_k$ achieves orthogonality of the columns in $\mGamma_k$. We therefore initialize diagonal precisions to $\mE_k = c\mI$ and diagonalize $\mM_k$ after each SGD step. This will ensure orthogonality of the columns in the loading matrix in the initial phases of SGD. 
As the $\mE_k$ evolve under the influence of SGD, loading matrix columns will no longer be orthogonal but still linearly independent by virtue of diagonalizing the $\mM_k$, see \cref{sec:prec}. 
\\
\par\noindent\textbf{Positive-definiteness of the} $\mM_k$
If this were not the case, the logarithm in \cref{eqn:precdet} would be undefined. 
The $\mE_k$ must be positive-definite in any case, which we achieve by re-parameterizing them as the square of a diagonal matrix: $\mE_k = \mathcal E_k^2$. Thus, the matrices $\mP_k$ are positive-definite as well.
For maintaining positive-definiteness of $\mM_k$, we must ensure that $\det M_k > 0$, which is not guaranteed, see \cref{sec:m}. Since the $\mM_k$ are diagonalized after each SGD step (see previous paragraph), their eigenvalues can be read off from their diagonal. Maintaining positive-definiteness then amounts to preventing negative eigenvalues during SGD.
Where an eigenvalue $(\mM_k)_{ii}$ is below a threshold $M_\text{min}$, we multiply the corresponding vector $\vGamma_{k:i}$ in the loading matrix $\mGamma_k$ by a factor that ensures that $(\mM_k)_{ii}=\vGamma_{k:i}^T\mE^{-1}\vGamma_{k:i}=M_\text{min}$. Multiplication by a constant factor is the only simple operation that
preserves the diagonality of the $\mM_k$ and the linear independence of the columns of the loading matrices. 
%
\section{Experiments}\label{sec:exp}
We created a publicly available implementation of SGD-based MFA training based on Tensorflow 2.7 \cite{abadi2016tensorflow}, in particular its keras package. 
This implementation is used for all experiments described here, with acceleration provided GPUs of the type "nVidia GTX 2080 Super".
Each experiment for which metrics are presented is repeated 10 times, and we always report means and standard deviations of these metrics. 

Unless otherwise stated, we always use $K=25$ components and a latent dimension of $l=4$. Otherwise, default parameters and initialization from \cite{gepperth2021c} are used. 
In particular, centroids are always initialized to uniform random values between -0.1 and 0.1. The diagonal precision matrices $\mathcal E_k$ are clipped from above at 
at value of $D_\text{max}=20$ in order to avoid unbounded precision values for pixels that have no variability. Throughout all experiments, a mini-batch size of 100 is used for SGD. 

In order to avoid undesirable local optima during early phases of training, it is imperative that centroids $\vmu_k$ converge before the precision and loading matrices $\mE_k$, $\mGamma_k$. In practice, this can be achieved by giving different weights to the gradients $\vec \nabla_{\vmu_k}\mathcal L$, $\vec \nabla_{\mE_k}\mathcal L$ and $\vec \nabla_{\mGamma_k}\mathcal L$. Sensible values for these weights are $\lambda_{\mGamma} = \lambda_{\mE} = 0.1$ and $\lambda_{\vmu}= 1$, although in this case the precision matrices will converge rather slowly. A workaround to artificially accelerate training is to conduct separate training phases: in phase I, only the $\vmu_k$ are adapted, whereas all quantities are adapted together with equal weights in phase II. Unless otherwise stated, we use 15 epochs for phase 1 and 50 epochs for phase II. 
\subsection{Training procedure and basic feasibility}\label{sex:exp1}
\begin{figure*}[t]
\centering
\includegraphics[width=0.16\textwidth]{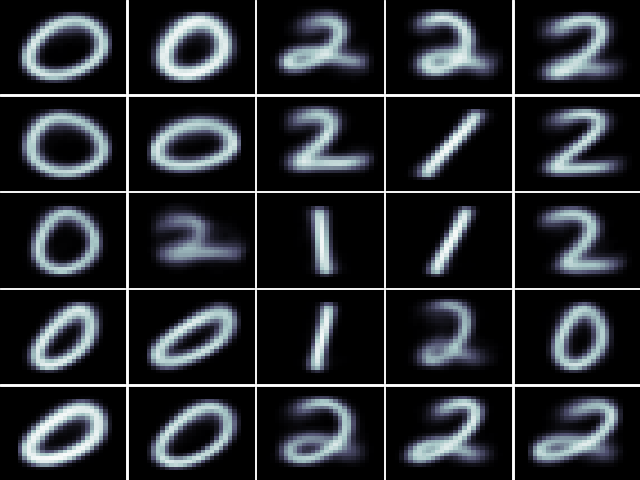}
\includegraphics[width=0.16\textwidth]{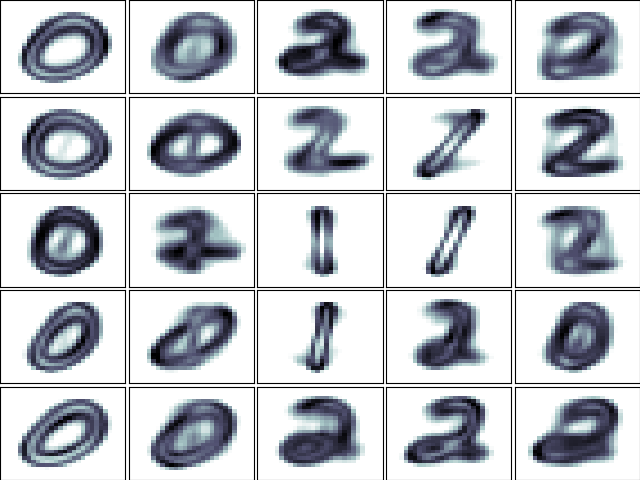}
\includegraphics[width=0.16\textwidth]{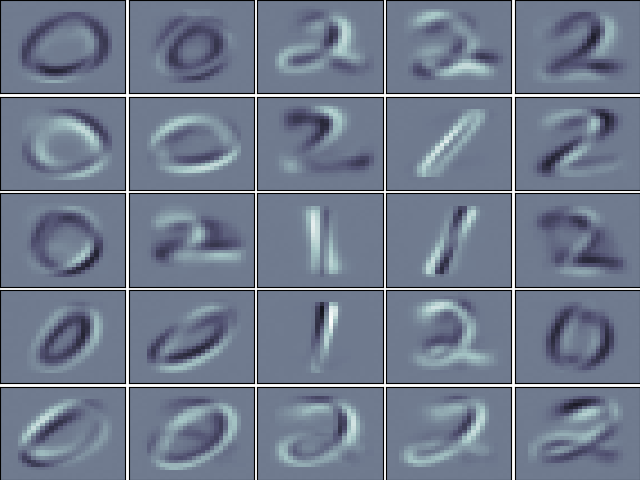}
\includegraphics[width=0.16\textwidth]{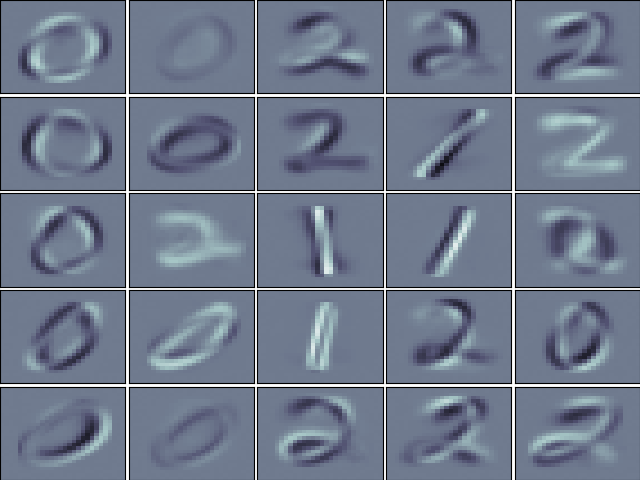}
\includegraphics[width=0.16\textwidth]{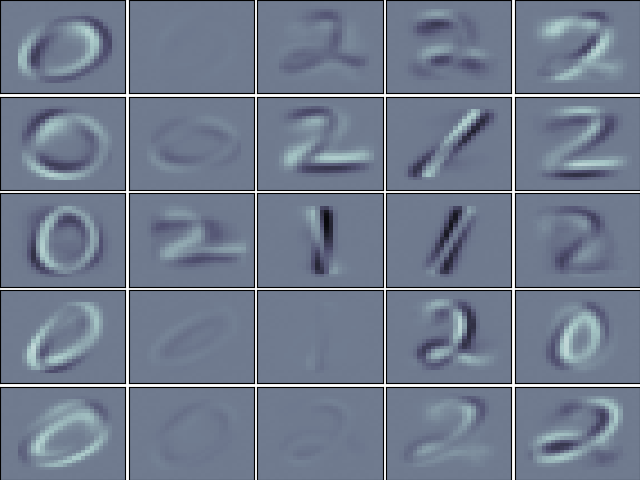}
\includegraphics[width=0.16\textwidth]{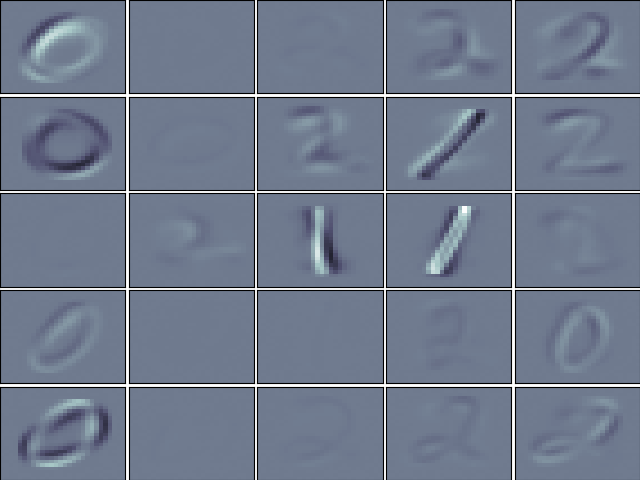} \\
\vspace{0.5cm}
\includegraphics[width=0.16\textwidth]{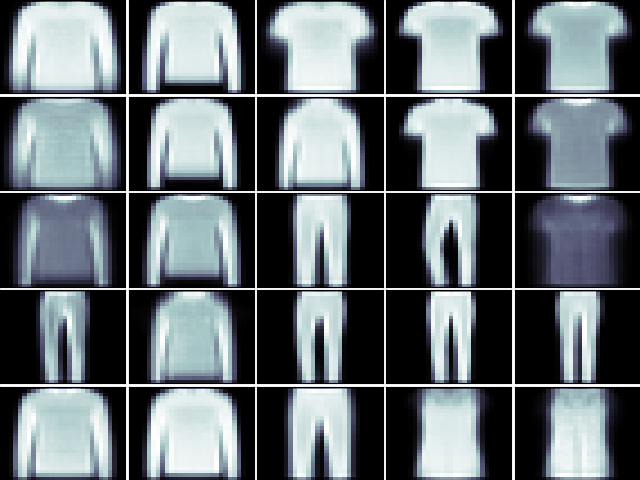}
\includegraphics[width=0.16\textwidth]{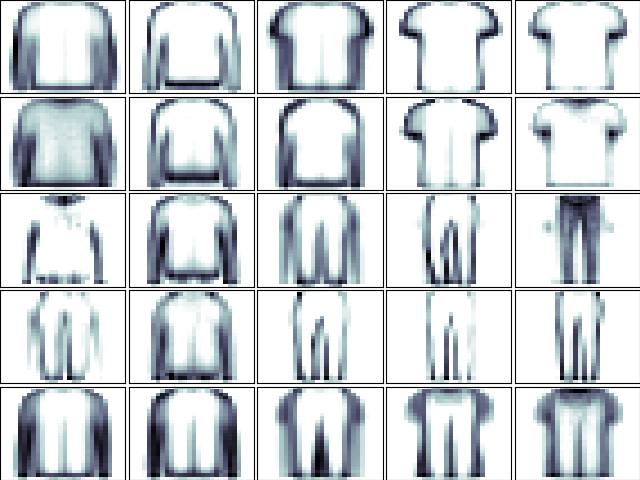}
\includegraphics[width=0.16\textwidth]{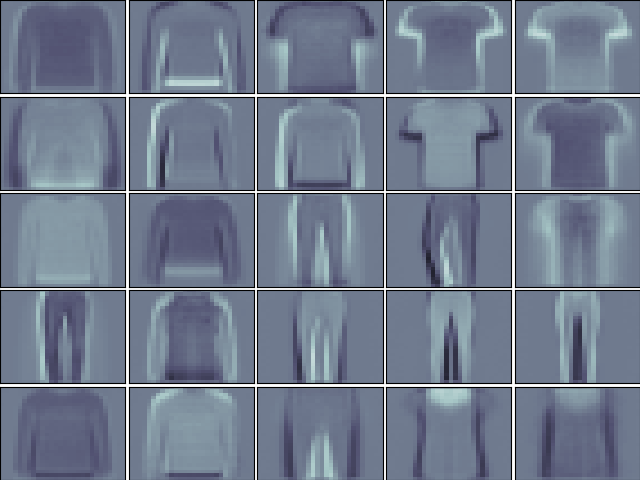}
\includegraphics[width=0.16\textwidth]{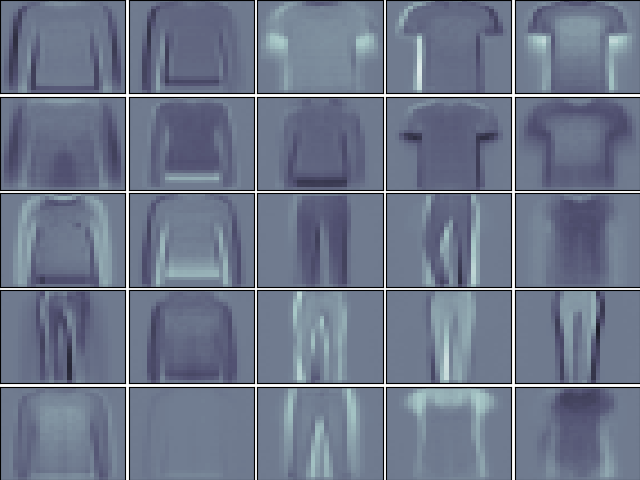}
\includegraphics[width=0.16\textwidth]{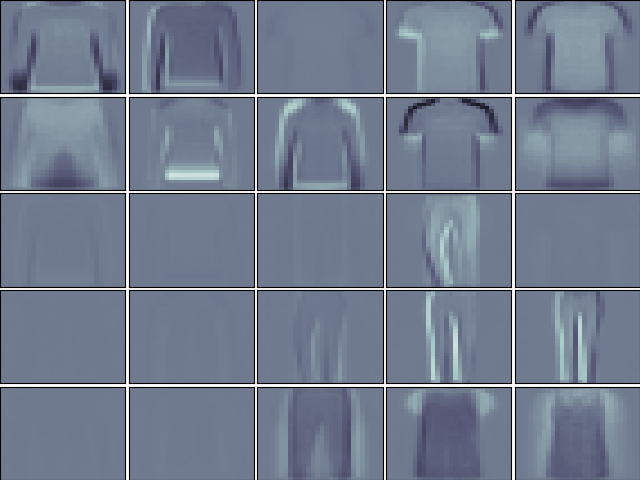}
\includegraphics[width=0.16\textwidth]{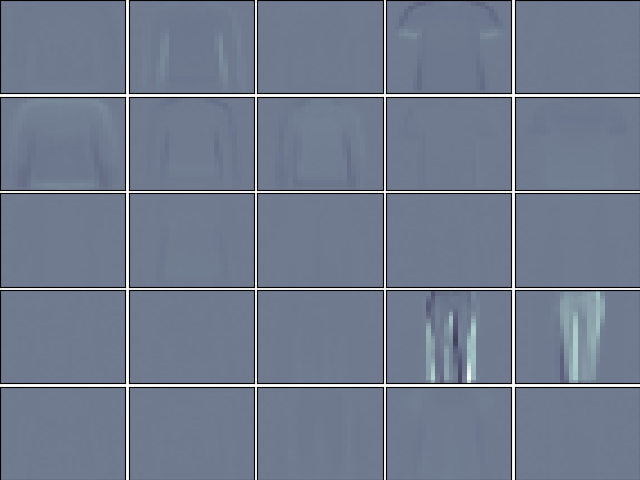}

\caption{\label{fig:res:m} Centroids $\vmu_k$ (left), precisions $\mE_k$ (second from left) and loading matrices $\mGamma_k$ for $l=0,1,2,3$ (four right-most images) for precision-based MFA performed on MNIST (upper row) and FashionMNIST(lower row). Each tile in the loading matrix images belongs to the centroid at the corresponding tile position. Centroids are scaled in the $[0,1]$-range, loading matrices in the $[-1,1]$-range and precisions between 18 and 20.
}
\end{figure*}

In order to demonstrate the feasibility of MFA training by constrained SGD, we train the MFA model as in \cref{sec:exp} on MNIST and FashionMNIST. To avoid cluttered and over-complex results, we restrict training to classes 0,1 and 2 (although it works just fine with all classes). For both datasets, we report the final centroids, precisions and loading matrices in \cref{fig:res:m}. As stated in \cite{gepperth2021c}, the centroids are initialized to random values between-0.1 and 0.1, variances are uniformly initialized to $D_\text{max}=20$ and loading matrices are initialized such that $\mM_k=0.0001 \mI$. The component weights are uniformly initialized to $\pi_k=\frac{1}{4}$. 

When repeating this experiment 10 times, we observe that convergence is always achieved, although of course the precise converged values vary due to random initial conditions. When inspecting the centroids, the SOM-like self-organization by similarity is apparent, which is an artifact of the training process, see \cref{fig:res:m}. 

It is visually apparent from \cref{fig:res:m} that the columns of the loading matrices are all distinct (due to diagonalizing $\mM_k$), and capture major directions of variations. We also note that the strength of variations decreases with higher $l$, reminiscent of principal directions in PCA. 
\subsection{MFA sampling}\label{sex:exp2}
\begin{figure*}
\centering
\includegraphics[width=0.32\textwidth]{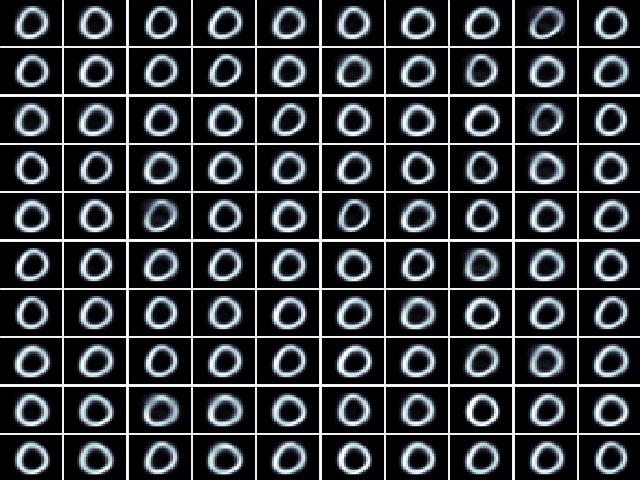}
\includegraphics[width=0.32\textwidth]{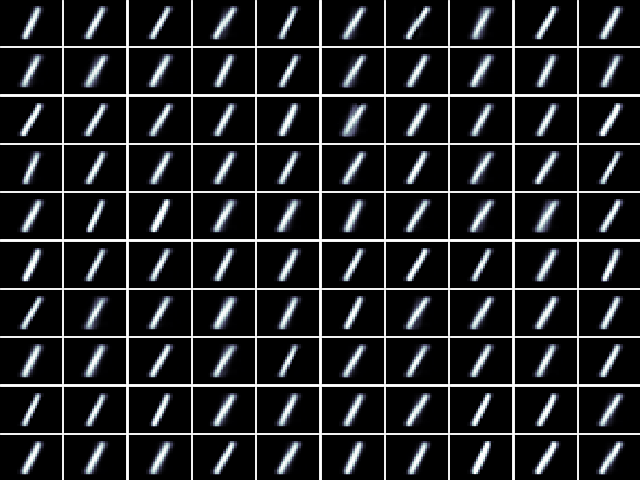}
\includegraphics[width=0.32\textwidth]{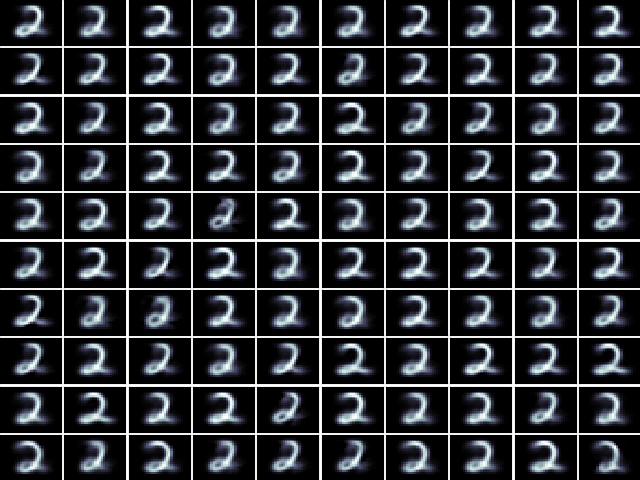} \\
\vspace{0.2cm}
\includegraphics[width=0.32\textwidth]{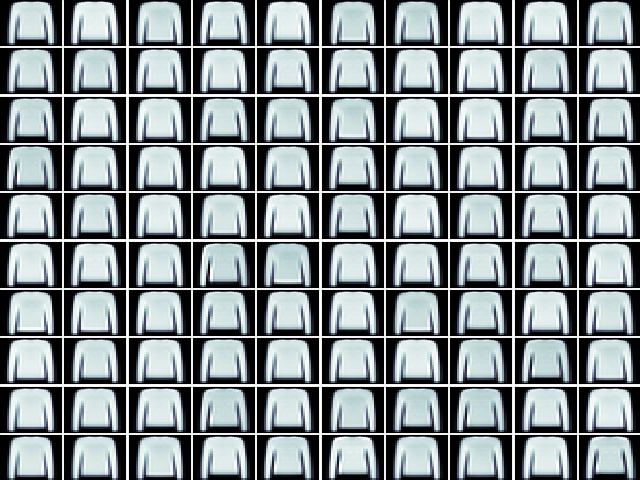}
\includegraphics[width=0.32\textwidth]{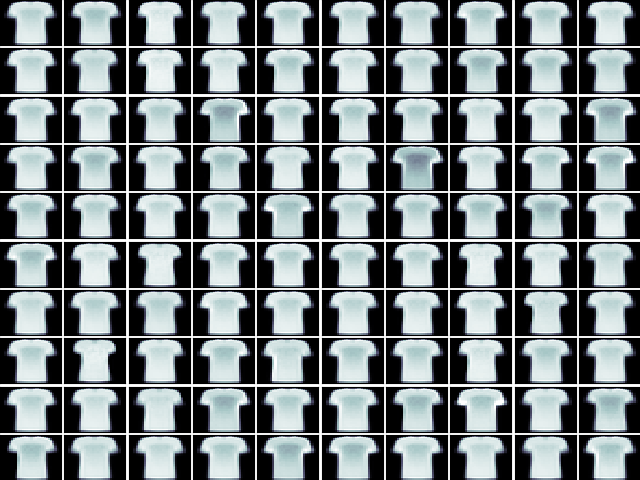}
\includegraphics[width=0.32\textwidth]{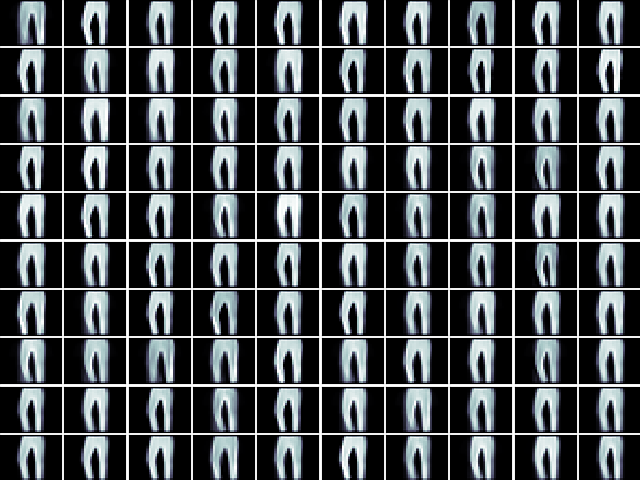} 
\caption{\label{fig:sampling}
Sampling from MNIST(upper row) and FashionMNIST(lower row) from selected mixture components. Selected components (to be compared to centroids in \cref{fig:res:m}) are 0,6,15 for MNIST and 1,3,13 for FashionMNIST. 
Please enlarge the figure in order to observe that each sample is distinct in shape from the others. This is most notable for the rightmost MNIST samples of digit class 2. But also for other classes, differences in slant and strokes are observable.
}
\end{figure*}
We use the trained models from \cref{sex:exp1} to perform sampling (see \cref{sec:sampling}) from mixture components 0,6 and 15 (MNIST) and 1,3,13 (FashionMNIST). As described in 
\cref{sec:sampling}, the mixture component to sample is be chosen randomly according to the component weights, but here we manually select components to sample from, in order
to illustrate how MFA introduces variability into sampling from the same component. 
Sampling results are presented in \cref{fig:sampling}. We observe that, despite the fact of originating from the same centroid, all samples are subtly different due to the multi-dimensional latent space that captures variations along the directions stored in the factor loadings. 
\subsection{Outlier detection and comparison to GMMs}
\begin{table}[t]
\centering
\begin{tabular}{ccc}
dataset & GMM & MFA \\
\hline
MNIST & 90.1 & 92.2 \\
FashionMNIST & 83.7 & 85.3 
\end{tabular}
\caption{\label{tab:auc}
AUC measures computed for outlier detection by precision-based MFA and vanilla GMM. Datasets are MNIST and Fashion MNIST, where classes 0-8 were used for training.
}
\end{table}
For demonstrating the outlier detection capacity and thus the validity of precision-based MFA, we train it on MNIST classes 0-8 and record the test test log-likelihoods on the remaining class 9. For this experiment, we use $K=49$ components but leave the experimental procedure and parameters of \cref{sec:exp} untouched otherwise. Since class 9 has not been used for training, it constitutes an outlier class and should be recognized as such. This is realized by a threshold $\theta$ applied to each log-likelihood $\mathcal L(\vx_n)$, where $\mathcal L(\vx_n) < \theta$ indicates an outlier since log-likelihoods are maximized. 

For a given value of $\theta$, we can compose two values: the percentage  $\alpha$ of true inliers (class 0-8 samples) that are recognized, and the percentage $\beta$ of true outliers (class 9 samples) that are rejected. 
By varying $\theta$, we obtain the ROC-like plots, for which we compute the area-under-the-curve measure (AUC). AUCs for MNIST and FashionMNIST, both for a GMM (same number of components) and precision-based MFA, are given in \cref{tab:auc}. We observe that MFA slightly improves upon GMM performance.
\section{Discussion}\label{sec:discussion} 
The mathematical and experimental results presented in \cref{sec:math} and \cref {sec:exp} indicate that, 
first of all, precision-based MFA can be successfully trained by SGD, from random initial conditions, and on large-scale datasets. We showed that MFA models trained by SGD can be successfully used for sampling and outlier detection, two important functionalities of unsupervised learning. Here, we will discuss the wider implications of these results:

\par\smallskip\noindent 
\textbf{Efficiency for streaming data} In conventional machine learning settings where a model is first trained, then deployed/applied without further adaptation, it is a feasible strategy to pre-compute the inverse covariance matrix after training has finished, and thus to avoid matrix inversions. 
In situations where the model is updated continuously while being applied (evaluate-then-train strategy, see \cite{gepperth2016tut}), this is no longer feasible since the covariance matrix will change continuously. Here, the precision-based approach will be superior since it requires matrix inversion for the training step only. 
\par\smallskip\noindent 
\textbf{Simplicity} 
The proposed SGD approach to MFA has the advantage of being extremely simple. In particular, no complex initialization of centroids by, e.g., k-means, is required, and neither is the even more complex initialization of covariance matrices as,e.g., used in \cite{Richardson2018}. Instead, centroids are initialized to random small values with guaranteed convergence, as shown here and in \cite{gepperth2021c}. 
\par\smallskip\noindent 
\textbf{Processing of large and high-dimensional datasets} Due to the "trick" of computing determinants and matrix inverses on the low-dimensional matrices $\mM_k$, $\mL_k$, MFA can be applied to high-dimensional data, at least as long as $l$ is small. This has been shown in previous works \cite{Ghahramani1997,McLachlan2003} using EM for optimization. However, MFA training on large datasets or streaming data has been problematic due to the batch-type nature of EM, which causes memory requirements to grow linearly with the number of samples. Stochastic variants of EM only partially fix this problem since they introduce several new and unintuitive hyper-parameters that must be tuned by grid search. Here, SGD offers a principled alternative, since its memory requirements only depend on the chosen mini-batch size, and since the choice of the single learning-rate parameter is well-understood.

\par\smallskip\noindent 
\textbf{Independent factor loadings} In contrast to the original formulation of MFA \cite{Ghahramani1997} which imposes no constraints on the loading matrices, we demand that factor loadings be independent. We do not observe any problems related to this additional constraint, as convergence was universal in all conducted experiments. Conversely, we did observe a few cases where SGD was stuck in local extremal points with partially identical factor loadings. 

\par\smallskip\noindent 
\textbf{Small-$l$ regime} 
MFA is strongly related to principal components analysis (PCA), since the factor loadings aim to capture, for each mixture component, the directions that best explain that component's variance. In contrast to PCA, we do not require individual directions to be orthogonal, only independent. As well-known fact is that the number of directions required to explain a large part of the variance is rather small, especially for images. Thus, running MFA in a "small $l$ regime" seems feasible.

\section{Conclusion and outlook}\label{sec:conclusion}
This article has a mathematical as well as practical contribution, showing how precision-based MFA can be trained by SGD in theory, and then validating the mathematical proofs by experiments using an own keras-based implementation. Future work will include a convolutional generalization of MFA, and the stacking into deep MFA hierarchies for realistic sampling od complex images.

\bibliography{iclr2021_conference}
\bibliographystyle{unsrt}

\end{document}